\documentclass[letterpaper, 10 pt, conference]{ieeeconf}  

\IEEEoverridecommandlockouts                              

\title{\LARGE \bf
TrajEdit: Real-Time Flow-Matching Trajectory Adaptation via FlowEdit and CBF Safety Filter} 
\author{Wondmgezahu Teshome, Arya Honarpisheh, Octavia Camps, Mario Sznaier%
\thanks{This work was supported in part by NSF grant CMMI--2208182 and ONR grant N00014-21-1-2431. The authors are with the Department of Electrical and Computer Engineering, Northeastern University, Boston, MA, USA.}}

\usepackage{amsmath}      
\usepackage{amssymb}      
\usepackage{amsfonts}     
\usepackage{mathtools}    
\usepackage{bm}           
\usepackage{newtxtext,newtxmath}

\usepackage{graphicx}     
\usepackage{placeins}
\usepackage{subfigure}    
\usepackage{threeparttable}
\usepackage{tikz}         
\usetikzlibrary{arrows.meta, positioning, shapes.geometric}

\newcommand{\bestm}[1]{\mathbf{#1}}
\newcommand{\secondm}[1]{\underline{#1}}

\usepackage{algorithm}
\usepackage{algpseudocode}
\usepackage{etoolbox}
\makeatletter
\patchcmd{\ALG@doentity}{\noindent\hskip\ALG@tlm}{%
  \parshape 2
    \@totalleftmargin \linewidth
    \dimexpr\@totalleftmargin+\ALG@tlm\relax
    \dimexpr\linewidth-\ALG@tlm\relax
  \noindent\hskip\ALG@tlm
}{}{\PackageError{TrajEdit}{Algorithm indentation patch failed}{Check the algorithmicx package version.}}
\makeatother

\usepackage{booktabs}     
\usepackage{multirow}     
\usepackage{array}        
\usepackage{cite}
\usepackage[colorlinks=true, citecolor=green, urlcolor=blue, linkcolor=blue, hypertexnames=false]{hyperref}
\usepackage{comment}
\usepackage{caption} 
\usepackage{xcolor}
\usepackage{colortbl}

\usepackage{booktabs}     
\usepackage{multirow}     
\usepackage{array}  
\begin{document}

\maketitle
\thispagestyle{empty}
\pagestyle{empty}

\begin{abstract}
Safe, real-time motion planning is especially difficult when a dynamic obstacle actively pursues the ego agent, requiring rapid adaptation as the environment changes. We present TrajEdit, a hierarchical trajectory planner for pursuit-evasion that combines flow matching, trajectory editing, and safety filtering. Rather than regenerating trajectories from scratch at every step, TrajEdit follows the current plan under nominal conditions and adapts its remaining portion through FlowEdit as the pursuer approaches, and falls back to a control barrier function safety filter if the adversarial agent is within a safety threshold. Our use of FlowEdit for trajectory editing reduces computation time while maintaining a comparable success rate. We evaluate TrajEdit in 2D pursuit-evasion with static obstacles and an adversarial pursuer, demonstrating high success rates with low capture and collision rates, while achieving computational time savings over full replanning baselines.
\end{abstract}

\begin{figure*}[t]
  \centering
  \includegraphics[trim={0cm 0cm 0cm 0cm}, clip,width=0.98\textwidth]{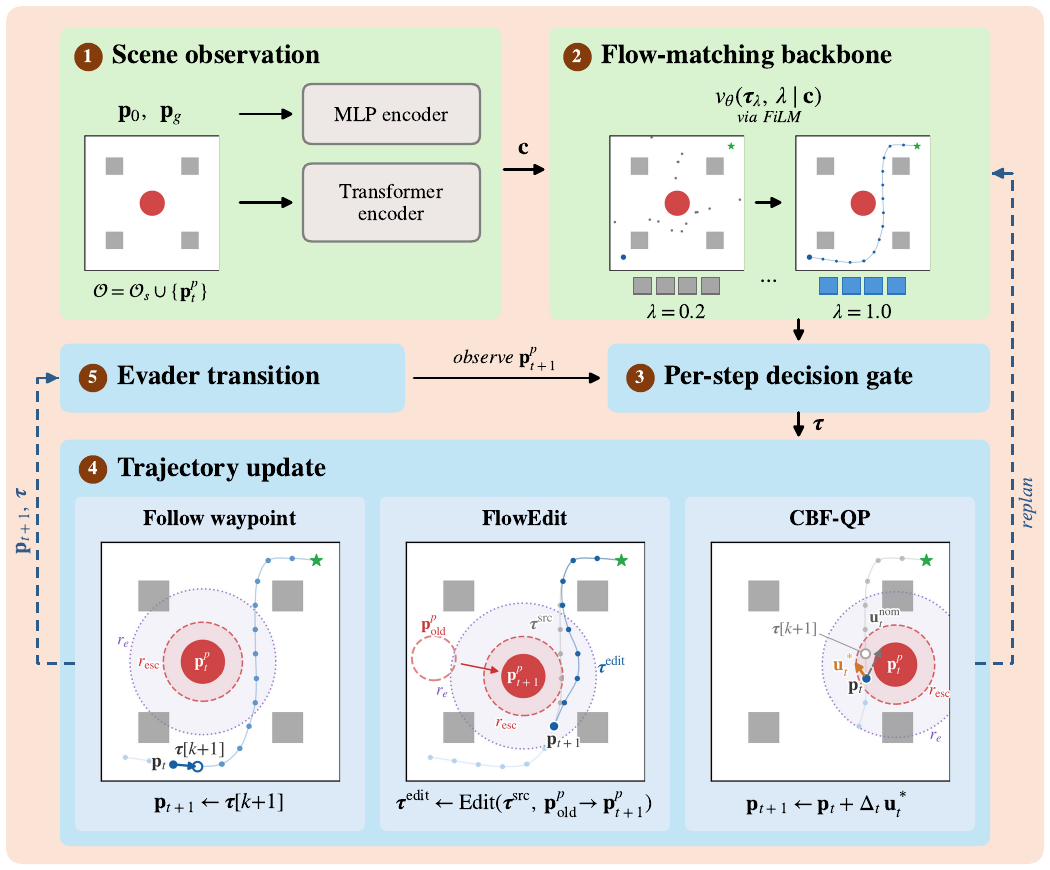} 
\caption{Overview of TrajEdit. (1)~The scene and plan endpoints define the condition $\mathbf{c}$. (2)~The flow-matching backbone integrates $v_\theta$ from noise to a plan $\boldsymbol{\tau}$. (3)--(5)~At each step, the CBF filter acts if no plan is accepted or if the evader or its next waypoint is within $r_{\mathrm{esc}}$ of the pursuer, and the plan is then regenerated. Otherwise, the evader follows the plan, which is kept while the pursuer stays at least $r_e$ away and enough of the plan remains, and edited if not. In (4), dark blue marks the remaining plan, light blue its executed part, and grey a discarded plan; dashed arrows indicate feedback, $\boldsymbol{\tau}[k{+}1]$ is the next waypoint of $\boldsymbol{\tau}$.}
  \label{fig:TrajEdit}
\end{figure*}

\section{Introduction}
\label{sec:ah_introduction}

Autonomous robots navigating changing environments must reach their goals while avoiding moving obstacles~\cite{paden2016survey}. Pursuit-evasion adds an adversarial element, as the pursuer actively attempts to capture the robot~\cite{chung2011search}. The pursuer's motion can make an existing trajectory unsafe, requiring timely responses to new observations. The planner must therefore balance responsiveness with the computational cost of updating its trajectory.

Classical motion planners address these requirements through geometric search and local obstacle avoidance. RRT-Connect~\cite{kuffner2000rrt} searches for a feasible path by growing trees from the start and goal. RRTX~\cite{otte2016rrtx} maintains and repairs a search graph as obstacles change. Artificial potential fields combine goal attraction with obstacle repulsion~\cite{khatib1986real}. Conventional formulations can encounter local minima, and extensions use virtual targets to restore progress toward the goal and modified repulsive fields to improve obstacle avoidance~\cite{zhai2024local}. The velocity obstacle (VO) method~\cite{fiorini1998motion} uses relative positions and velocities to identify motions that would lead to collision and select an avoidance velocity. The computational demands of online search and the local minima of potential fields motivate learning motion policies and trajectory priors from data.

Learning-based methods have been applied to both action selection and trajectory generation. Reinforcement learning (RL) has been used to learn evasive actions from visual observations~\cite{zhu2018learning} and has been combined with imitation learning for goal-directed escape among obstacles~\cite{qu2023pursuit}. Generative models learn distributions over trajectories. Diffuser~\cite{janner2022planning} generates plans through iterative denoising, and Motion Planning Diffusion~\cite{carvalho2023motion} combines learned trajectory priors with guidance from planning costs. In pursuit-evasion, diffusion has also been combined with a local RL policy to coordinate global planning and evasive behavior~\cite{wu2024diffusion}. Repeated denoising can become costly when plans must be updated frequently. Flow matching~\cite{lipmanflow} offers an alternative by learning a continuous velocity field and generating samples through an ordinary differential equation. FlowMP~\cite{nguyen2025flowmp} applies this approach to robot motion planning and reports lower inference times than diffusion baselines in its evaluated settings.

An additional strategy is to update the remaining portion of the active plan as new observations arrive, rather than regenerate the entire trajectory from scratch. Warm starts allow a diffusion planner to perturb and refine its previous trajectory under updated observations~\cite{beyer2025joint}. RAMP~\cite{teshome2025real} uses limited denoising and artificial potential fields to refine trajectories in pursuit-evasion. Adaptive online replanning~\cite{zhou2023adaptive} selects between retaining a plan, repairing its remaining portion, and generating a replacement. Other approaches distribute denoising across execution steps, either by refining a plan with increasing noise levels along its horizon~\cite{guo2026efficient} or by maintaining a queue of partially denoised actions with feedback guidance~\cite{ye2025ra}. These methods establish the value of trajectory reuse and incremental updates. Our focus is on how conditional flow editing can update the remaining trajectory as the observed scene changes.

For this purpose, we adapt FlowEdit~\cite{kulikov2025flowedit}, originally proposed for image editing, to trajectory adaptation. FlowEdit constructs an editing path using differences between learned velocity fields under source and target conditions, without first inverting the source sample to noise. Related work has applied conditional flow editing to human motion under changes in text or skeleton conditions~\cite{li2026unified}. Here, the conditions change as the ego agent and pursuer move. We use the previous and current conditions to edit the unexecuted portion of the plan while retaining its goal. The same learned flow model supports initial generation and subsequent editing, allowing the planner to update an existing trajectory without generating a new one from noise at every step.

Unlike image editing, this application also requires consideration of safety. Control barrier functions (CBFs) provide a way to constrain control inputs using a specified safe set~\cite{ames2016control}. Recent work has incorporated these constraints into flow matching. SafeFlow~\cite{dai2025safeflow} adds barrier corrections during trajectory generation, while SafeFlowMatcher~\cite{yang2026safeflowmatcher} first predicts a path and then refines it through a separate correction phase. In both methods, barrier corrections act within the trajectory-generation process, and their navigation evaluations use static obstacles. In pursuit-evasion, the planner must update its plan as the pursuer moves while responding to immediate capture risk during execution.

To coordinate plan updates with immediate evasion, we propose TrajEdit, a hierarchical planner for pursuit-evasion. Under nominal conditions, the ego agent follows the current trajectory. As the pursuer approaches, FlowEdit adapts the remaining plan to the new observation. When immediate evasion is needed, the CBF safety filter intervenes and the planner attempts to generate a new trajectory from the updated state. Our contributions can be summarized as follows:
\begin{itemize}
    \item A unified framework for pursuit-evasion that adapts FlowEdit to online trajectory editing and integrates CBF safety filtering during execution. Selective reuse and editing of the remaining trajectory avoid full regeneration at every update, addressing computational efficiency alongside safety.
    \item A systematic evaluation of TrajEdit against existing planning methods, together with ablation studies of its key components, to assess both task performance and computational efficiency.
\end{itemize}
\section{Problem Formulation}
\label{sec:ah_problem}

We consider an ego agent, referred to as the evader, that seeks to reach a goal region while avoiding static obstacles and an adversarial pursuer. Let $\mathcal{W}\subset\mathbb{R}^{2}$ denote the workspace, $\mathcal{O}_s=\{O_1,\ldots,O_m\}$ the static obstacle components, with $O_i\subset\mathcal{W}$, and $\mathcal{G}\subset\mathcal{W}\setminus\bigcup_{i=1}^{m}O_i$ the goal region. At step $t$, the evader's state is $\mathbf{s}_t=(\mathbf{p}_t,\mathbf{v}_t)\in\mathcal{S}\subset\mathbb{R}^{4}$, with position $\mathbf{p}_t$ and velocity $\mathbf{v}_t$. For sampling interval $\Delta_t>0$ and speed bound $v_{\max}$, its motion satisfies $\|\mathbf{p}_{t+1}-\mathbf{p}_t\|_2\leq v_{\max}\Delta_t$. The evader observes $\mathbf{o}_t=(\mathbf{s}_t,\mathbf{p}_t^{p},\mathcal{O}_s)$, where $\mathbf{p}_t^{p}\in\mathcal{W}$ is the pursuer's current position. The pursuer's policy and future positions are unknown, but its speed bound $v_{\max}^{p}$ is known, giving $\|\mathbf{p}_{t+1}^{p}-\mathbf{p}_t^{p}\|_2\leq v_{\max}^{p}\Delta_t$. Capture occurs when $\|\mathbf{p}_t-\mathbf{p}_t^{p}\|_2<r_{\mathrm{cap}}$, where $r_{\mathrm{cap}}>0$ is the Euclidean capture radius, and static collision when $\mathbf{p}_t\in O_i$ for some $i$.

A candidate trajectory is $\boldsymbol{\tau}=(\mathbf{s}_{t\mid t},\ldots,\mathbf{s}_{t+H-1\mid t})\in\mathbb{R}^{H\times 4}$, where $H$ is the planning horizon, and $\mathbf{s}_{t+k\mid t}$ is the state planned for step $t+k$ using information available at $t$. The plan starts at the evader's current position and may be updated as new observations arrive. We choose a target position $\mathbf{p}_g\in\mathcal{G}$ and use the planned goal state $\mathbf{s}_{\mathrm{goal}}=(\mathbf{p}_g,\mathbf{0})$.

Given the initial state $\mathbf{s}_0$ and the goal state $\mathbf{s}_{\mathrm{goal}}$, the objective is an online planning policy that respects the velocity bounds and reaches the goal at a finite step $T$ without collision or capture:
\begin{equation}
\label{eq:ah_planning_objective}
\begin{aligned}
    &\mathbf{p}_T\in\mathcal{G},\\
    &\mathbf{p}_t\in\mathcal{W}\setminus\bigcup_{i=1}^{m}O_i,\quad
    \|\mathbf{p}_t-\mathbf{p}_t^{p}\|_2\geq r_{\mathrm{cap}},\quad \forall\,t=0,\ldots,T.
\end{aligned}
\end{equation}

\section{TrajEdit} 
\label{sec:method}
We present TrajEdit, a hierarchical trajectory planner for pursuit-evasion: a flow matching model generates a nominal trajectory, FlowEdit adapts its remaining portion as the pursuer moves, and a CBF filter intervenes when immediate evasion is needed. At each step, the evader uses the current observation and remaining plan to choose between following the trajectory and applying the CBF filter. After both agents move, a fresh pursuer observation is available for planning. Under nominal conditions, the remaining plan is retained without adaptation. As the pursuer approaches, FlowEdit updates the remaining plan instead of regenerating it from scratch to reduce the computational cost of adaptation. After a filter intervention, the planner attempts to generate a new trajectory from the resulting evader position and updated pursuer observation. The complete planning algorithm is presented in Algorithm~\ref{alg:TrajEdit}. The following subsections detail each component: trajectory generation using flow matching (\S\ref{sec:flow_gen}), trajectory adaptation via FlowEdit (\S\ref{sec:flowedit_replan}), and CBF safety filtering (\S\ref{sec:cbf_safety}).

\begin{algorithm}[t]
\caption{TrajEdit: adaptive pursuit-evasion planning}
\label{alg:TrajEdit}
\begin{algorithmic}[1]
\State \textbf{Input:} Learned velocity field $v_\theta$, initial position $\mathbf{p}_0$, goal $\mathbf{s}_{\mathrm{goal}}$, obstacles $\mathcal{O}_s$, radii $r_e,r_{\mathrm{esc}}$.
\State $t\gets0$. Observe $\mathbf{p}_0^p$ and set $\mathbf{p}_{\mathrm{old}}^p\gets\mathbf{p}_0^p$.
\State Generate $\boldsymbol{\tau}$ with $v_\theta$ conditioned on $((\mathbf{p}_0,\mathbf{0}),\mathbf{s}_{\mathrm{goal}},\mathcal{O}_s,\mathbf{p}_0^p)$.
\While{not terminated}
    \If{$\boldsymbol{\tau}=\varnothing$ or either $\mathbf{p}_t$ or next waypoint is within $r_{\mathrm{esc}}$ of $\mathbf{p}_t^p$}
        \State Obtain $\mathbf{u}_t^*$ from~\eqref{eq:cbf_qp}, or terminate if infeasible
        \State Execute $\mathbf{p}_{t+1}\gets\mathbf{p}_t+\Delta_t\mathbf{u}_t^*$ and observe $\mathbf{p}_{t+1}^p$
        \State Regenerate $\boldsymbol{\tau}$ from $\mathbf{p}_{t+1}$ using $\mathbf{p}_{t+1}^p$
    \Else
        \State Follow one step to $\mathbf{p}_{t+1}$ and observe $\mathbf{p}_{t+1}^p$
        \If{$\|\mathbf{p}_{t+1}-\mathbf{p}_{t+1}^p\|_2<r_e$}
            \State Form $\boldsymbol{\tau}^{\mathrm{src}}$, $\mathbf{c}_{\mathrm{src}}$, and $\mathbf{c}_{\mathrm{tar}}$ as in~\S\ref{sec:flowedit_replan}
            \State Update $\boldsymbol{\tau}$ using Algorithm~\ref{alg:edit}
        \EndIf
    \EndIf
    \State On accepting a new plan, set $\mathbf{p}_{\mathrm{old}}^p\gets\mathbf{p}_{t+1}^p$
    \State $t\gets t+1$
\EndWhile
\end{algorithmic}
\end{algorithm}

\subsection{Conditional Flow Matching Trajectory Generation} \label{sec:flow_gen}
Following prior work on flow-matching motion planning~\cite{nguyen2025flowmp}, we use conditional flow matching~\cite{lipmanflow,tong2023improving} to generate nominal trajectories. We denote the starting condition of a plan by $\mathbf{s}_{\mathrm{start}}$, distinct from the episode's initial state $\mathbf{s}_0$. When generating a plan at step $t$, we set $\mathbf{s}_{\mathrm{start}}=(\mathbf{p}_t,\mathbf{0})$, using the current position and zero velocity as the planning boundary condition. The condition is $\mathbf{c}=(\mathbf{s}_{\mathrm{start}},\mathbf{s}_{\mathrm{goal}},\mathcal{O}_s,\mathbf{p}_t^p)$, containing the plan endpoints, static obstacles, and observed pursuer position.

Given a training trajectory $\boldsymbol{\tau}_1$, we sample Gaussian noise for the intermediate states of $\boldsymbol{\tau}_0$ while fixing both trajectories to the same endpoint states. The training path is the linear interpolant
\begin{equation}
    \label{eq:interp}
    \boldsymbol{\tau}_\lambda=(1-\lambda)\boldsymbol{\tau}_0
    +\lambda\boldsymbol{\tau}_1,
\end{equation}
where $\lambda\in[0,1]$ is generative time, distinct from the physical step $t$. The field $v_\theta(\boldsymbol{\tau}_\lambda,\lambda\mid\mathbf{c})\in\mathbb{R}^{H\times4}$ is trained to match the derivative of this path through
\begin{equation}
    \mathcal{L}_{\mathrm{CFM}}(\theta)=\mathbb{E}
    \left[\left\|v_\theta(\boldsymbol{\tau}_\lambda,\lambda\mid\mathbf{c})
    -(\boldsymbol{\tau}_1-\boldsymbol{\tau}_0)\right\|^2\right],
    \label{eq:cfm_loss}
\end{equation}
with expectation over training trajectories and their conditions, Gaussian noise, and $\lambda$ sampled uniformly from $[0,1]$.

During training, we randomly omit conditioning information so that the same network learns conditional and unconditional vector fields. At inference, classifier-free guidance~\cite{ho2022classifier} combines these fields as $v_\theta(\cdot\mid\mathbf{c};w)=(1-w)v_\theta(\cdot\mid\varnothing)+w\,v_\theta(\cdot\mid\mathbf{c})$, where $\varnothing$ denotes omitted conditioning and $w$ controls the influence of the conditioning information. A candidate is generated from fresh noise by integrating its intermediate states according to
\begin{equation}
    \frac{d\boldsymbol{\tau}_\lambda}{d\lambda}
    =v_\theta(\boldsymbol{\tau}_\lambda,\lambda\mid\mathbf{c};w),
    \qquad \lambda\in[0,1],
    \label{eq:ode_sampling}
\end{equation}
while preserving the endpoints. For generation and editing, candidates are screened for workspace and obstacle clearance, and against the pursuer while it is within $r_e$ of the evader. The candidate with greatest minimum clearance among those that pass is selected. If none passes, no plan is accepted. The resulting plan is reused and updated as new observations become available.

\begin{algorithm}[t]
\caption{FlowEdit trajectory adaptation}
\label{alg:edit}
\begin{algorithmic}[1]
\State \textbf{Input:} Learned velocity field $v_\theta$, source $\boldsymbol{\tau}^{\mathrm{src}}$, conditions $\mathbf{c}_{\mathrm{src}},\mathbf{c}_{\mathrm{tar}}$, steps $N,n_{\max}$, guidance scales $w_{\mathrm{src}},w_{\mathrm{tar}}$, candidates $N_c$.

\State $\Delta_\lambda\gets1/N$, $\lambda_0\gets1-n_{\max}/N$. Initialize $N_c$ copies of $\mathbf{z}^{\mathrm{edit}}_{\lambda_0}\gets\boldsymbol{\tau}^{\mathrm{src}}$.
\For{$i=N-n_{\max},\ldots,N-1$}
    \State $\lambda\gets i\Delta_\lambda$. Sample fresh $\boldsymbol{\epsilon}\sim\mathcal{N}(\mathbf{0},\mathbf{I})$ independently for each candidate.
    \State Construct $\mathbf{z}^{\mathrm{src}}_\lambda,\mathbf{z}^{\mathrm{tar}}_\lambda$ using~\eqref{eq:flowedit_inputs}.
    \State Compute $\mathbf{V}^{\Delta}_\lambda$ with guidance scales $w_{\mathrm{src}},w_{\mathrm{tar}}$ using~\eqref{eq:flowedit_direction}.
    \State $\mathbf{z}^{\mathrm{edit}}_{\lambda+\Delta_\lambda}\gets\mathbf{z}^{\mathrm{edit}}_\lambda+\Delta_\lambda\mathbf{V}^{\Delta}_\lambda$.
    \State Restore target start and goal states.
\EndFor
\State \Return the best trajectory, or failure.
\end{algorithmic}
\end{algorithm}

\subsection{Trajectory Adaptation via FlowEdit}
\label{sec:flowedit_replan}
As the environment evolves, the conditions under which the current plan was generated may no longer hold, and its unexecuted waypoints may become unsafe. Regenerating from noise discards a plan that may still be largely useful. Instead, we adapt it using FlowEdit~\cite{kulikov2025flowedit}, originally developed for text-conditioned image editing and subsequently applied by Li et al.~\cite{li2026unified} to human motion editing and retargeting. Here, we use the same editing principle to update the unexecuted trajectory under changing scene observations (Algorithm~\ref{alg:edit}).

When the CBF filter is inactive, the evader first follows the plan for one step. After both agents move, adaptation is triggered if $\|\mathbf{p}_{t+1}-\mathbf{p}_{t+1}^p\|_2<r_e$, where $r_e>0$ is the edit radius, or if the remaining plan has insufficient horizon. If too few states remain for editing, the planner regenerates instead. The remaining plan is retained when it has sufficient horizon and the updated separation is at least $r_e$.

Let $\boldsymbol{\tau}^{\mathrm{src}}$ denote the remaining portion of the current plan, called the source trajectory. Its first state, $\mathbf{s}_{\mathrm{start}}^{\mathrm{src}}$, retains the planned position and velocity, while $\mathbf{p}_{\mathrm{old}}^p$ denotes the pursuer observation associated with that plan, which may have been generated or last edited several steps earlier. After both agents move in Algorithm~\ref{alg:TrajEdit}, the target starting state is $\mathbf{s}_{\mathrm{start}}^{\mathrm{tar}}=(\mathbf{p}_{t+1},\mathbf{0})$, and $\mathbf{p}_{t+1}^p$ is the newly observed pursuer position. When the evader reaches the next waypoint, the source and target starting positions coincide, but the full states coincide only if the source velocity is zero. If the speed bound prevents reaching that waypoint, the positions may also differ. The source and target conditions are
\begin{equation}
\label{eq:edit_conditions}
\begin{aligned}
\mathbf{c}_{\mathrm{src}}&=(\mathbf{s}_{\mathrm{start}}^{\mathrm{src}},\mathbf{s}_{\mathrm{goal}},\mathcal{O}_s,\mathbf{p}_{\mathrm{old}}^p),\\
\mathbf{c}_{\mathrm{tar}}&=(\mathbf{s}_{\mathrm{start}}^{\mathrm{tar}},\mathbf{s}_{\mathrm{goal}},\mathcal{O}_s,\mathbf{p}_{t+1}^p).
\end{aligned}
\end{equation}
The goal and static obstacles are unchanged. Following the shared-noise construction of FlowEdit~\cite{kulikov2025flowedit} and its human-motion formulation~\cite{li2026unified}, we couple the source and target branches without inverting $\boldsymbol{\tau}^{\mathrm{src}}$ to noise.

Let $\mathbf{z}^{\mathrm{edit}}_\lambda$ denote the trajectory being edited at generative time $\lambda$, initialized at $\boldsymbol{\tau}^{\mathrm{src}}$. To evaluate the source- and target-conditioned fields, we construct auxiliary noisy trajectories $\mathbf{z}^{\mathrm{src}}_\lambda$ and $\mathbf{z}^{\mathrm{tar}}_\lambda$, respectively, using shared standard Gaussian noise $\boldsymbol{\epsilon}$ sampled at each step:
\begin{equation}
\label{eq:flowedit_inputs}
\begin{aligned}
    \mathbf{z}^{\mathrm{src}}_\lambda &= \lambda\,\boldsymbol{\tau}^{\mathrm{src}}
    + (1-\lambda)\,\boldsymbol{\epsilon},\\
    \mathbf{z}^{\mathrm{tar}}_\lambda &= \mathbf{z}^{\mathrm{edit}}_\lambda
    + \mathbf{z}^{\mathrm{src}}_\lambda - \boldsymbol{\tau}^{\mathrm{src}}.
\end{aligned}
\end{equation}
The source start and goal states are restored before constructing $\mathbf{z}^{\mathrm{tar}}_\lambda$, followed by the target start and goal states. The edit direction is the difference between the guided fields evaluated at these inputs,
\begin{equation}
\label{eq:flowedit_direction}
    \mathbf{V}^{\Delta}_\lambda
    =v_\theta\!(\mathbf{z}^{\mathrm{tar}}_\lambda,
    \lambda\mid\mathbf{c}_{\mathrm{tar}};w_{\mathrm{tar}})
    -v_\theta\!(\mathbf{z}^{\mathrm{src}}_\lambda,\lambda\mid
    \mathbf{c}_{\mathrm{src}};w_{\mathrm{src}}),
\end{equation}
where $v_\theta(\cdot\,;w)$ is the guided field defined in~\S\ref{sec:flow_gen}, and $w_{\mathrm{src}}$ and $w_{\mathrm{tar}}$ are the source and target guidance scales.
Integrating this difference from
$\mathbf{z}^{\mathrm{edit}}_{\lambda_0} = \boldsymbol{\tau}^{\mathrm{src}}$, with generative step size $\Delta_\lambda=1/N$,
\begin{equation}
    \mathbf{z}^{\mathrm{edit}}_{\lambda+\Delta_\lambda} = \mathbf{z}^{\mathrm{edit}}_{\lambda}
    + \Delta_\lambda \, \mathbf{V}^{\Delta}_\lambda ,
\end{equation}
updates the trajectory toward the new condition, with
the endpoint constraints and workspace bounds re-imposed after each step.
We use only the last $n_{\max}<N$ of the $N$ integration steps, starting at $\lambda_0=1-n_{\max}/N$. The final state $\mathbf{z}^{\mathrm{edit}}_1$ is the adapted trajectory, obtained by reusing the existing plan without inversion or full integration from noise. We form $N_c$ candidates using independent noise and apply the screening and selection described in~\S\ref{sec:flow_gen}.

FlowEdit was developed for conditional content editing, whereas our trajectories must also support collision avoidance during execution. Conditioning on the observed scene and restoring endpoint constraints do not by themselves enforce collision avoidance. We therefore complement trajectory adaptation with the CBF safety filter in~\S\ref{sec:cbf_safety}, which intervenes when the pursuer approaches.

\subsection{CBF Safety Filter}
\label{sec:cbf_safety}

We complement trajectory editing with a CBF filter~\cite{ames2016control} for immediate evasion during execution. It activates when either the current position or the next planned waypoint lies within the escape radius $r_{\mathrm{esc}}$ of the observed pursuer, or when no accepted plan is available. The filter selects a velocity $\mathbf{u}_t$ for the update $\mathbf{p}_{t+1}=\mathbf{p}_t+\Delta_t\,\mathbf{u}_t$. To bias the motion toward the goal, we use the reference velocity
\begin{equation}
\label{eq:cbf_nominal}
    \mathbf{u}_t^{\mathrm{nom}}
    =v_{\max}\frac{\mathbf{p}_g-\mathbf{p}_t}
    {\|\mathbf{p}_g-\mathbf{p}_t\|_2},
    \qquad \mathbf{p}_t\neq\mathbf{p}_g.
\end{equation}
We represent static obstacles as unions of box components $O_1,\ldots,O_m$. For each component $O_i$, let $\Pi_{O_i}(\mathbf{p}_t)$ be the Euclidean projection of $\mathbf{p}_t$ onto $O_i$, and define $\mathbf{d}_{i,t}=\mathbf{p}_t-\Pi_{O_i}(\mathbf{p}_t)$. With relative position $\mathbf{e}_t=\mathbf{p}_t-\mathbf{p}_t^p$, we use the barriers
\begin{equation}
\label{eq:cbf_barriers}
    b_{i,t}=\|\mathbf{d}_{i,t}\|_2^2-r_s^2,
    \qquad b_{p,t}=\|\mathbf{e}_t\|_2^2-\rho^2,
\end{equation}
where $r_s>0$ is the static clearance margin and $\rho>r_{\mathrm{cap}}$ is the pursuer separation radius, providing an additional margin beyond the Euclidean capture radius. Nonnegative barriers ensure clearance from static obstacles and separation from the pursuer.

Although the pursuer's velocity is unknown, its speed bound gives the worst-case bound term $2\|\mathbf{e}_t\|_2v_{\max}^p$. Including this term in the pursuer constraint, we define the filtered velocity $\mathbf{u}_t^*$ through the hard-constrained problem
\begin{equation}
\label{eq:cbf_qp}
\begin{aligned}
 \min_{\mathbf{u}}\quad
     &\tfrac12\|\mathbf{u}-\mathbf{u}_t^{\mathrm{nom}}\|_2^2\\
 \text{s.t.}\quad
     &2\mathbf{d}_{i,t}^\top\mathbf{u}\geq-\gamma b_{i,t},\quad i=1,\ldots,m,\\
     &2\mathbf{e}_t^\top\mathbf{u}
       \geq-\gamma b_{p,t}+2\|\mathbf{e}_t\|_2v_{\max}^p,\\
     &\|\mathbf{u}\|_2\leq v_{\max},\quad
       \mathbf{p}_t+\Delta_t\,\mathbf{u}\in\mathcal{W}.
\end{aligned}
\end{equation}
Here $\gamma>0$ is the barrier decay rate. After a feasible filter step, regeneration is attempted from the updated evader position using the new pursuer observation. Failed planning keeps the filter active, while failure to obtain an admissible velocity terminates the simulation before execution. Under exact observations and execution, with $\mathbf{p}_0\in\mathcal{W}$, initially nonnegative barriers, and $0<\gamma\Delta_t\leq1$, the constraints in~\eqref{eq:cbf_qp} imply the standard discrete-time barrier condition~\cite{zeng2021safety}
\begin{equation}
\label{eq:cbf_discrete_bound}
    b_{j,t+1}\geq(1-\gamma\Delta_t)b_{j,t}\geq0,
    \quad j\in\{1,\ldots,m,p\}.
\end{equation}
Thus, every feasible filter step preserves nonnegative barriers.

Safety must also be maintained while the evader follows a plan without filter intervention. Each plan starts at the evader's position when adopted, and its segments are checked to lie inside the workspace and maintain at least $r_s$ clearance from static obstacles. Following these segments within the speed limit preserves this clearance. When the filter is inactive, the current pursuer separation is at least $r_{\mathrm{esc}}$. Since the agents can close this gap by at most $(v_{\max}+v_{\max}^p)\Delta_t$ in one step, we choose
\begin{equation}
\label{eq:cbf_trigger_margin}
    r_{\mathrm{esc}}\geq\rho+(v_{\max}+v_{\max}^p)\Delta_t,
\end{equation}
so that separation is still at least $\rho$ at the next control decision. Under the assumptions above, this condition and~\eqref{eq:cbf_discrete_bound} ensure that every executed step ends without collision or capture.

\section{Experiments}
We evaluate TrajEdit on 2D pursuit-evasion tasks in which an evader must reach a goal while avoiding static obstacles and a pursuer.
\subsection{Experimental Setup}
\paragraph{Dataset and evaluation}
All experiments take place in the workspace $[-1,1]^2$ on two obstacle layouts. The \emph{convex} layout is a fixed configuration of four disjoint box obstacles with size $0.26$. In the \emph{concave} layout, each start-goal pair has its own arrangement of three L-shaped obstacles, each formed from three boxes. Following~\cite{carvalho2023motion}, training trajectories are generated using RRT-Connect~\cite{kuffner2000rrt}, B-spline interpolation, and GPMP optimization~\cite{mukadam2018continuous}. Each state consists of 2D position and velocity ($d=4$), and each trajectory has 48 timesteps. The convex training set consists of 2{,}000 configurations of six boxes, each with 20 start-goal pairs and up to 25 trajectories per pair. The concave training set consists of 500 configurations, each with 15 start-goal pairs and 10 trajectories per pair, and every arrangement contains ten boxes, nine forming the concave shape and one placed elsewhere. All test layouts are unseen during training. The obstacle encoder is trained on all boxes; at inference, the final slot is assigned to the pursuer, and any unused slot repeats an obstacle.

The convex layout is evaluated on 100 start-goal pairs, and the concave layout on 100 configurations, each with its own start-goal pair. Every pair is evaluated from five initial pursuer positions, and we report the mean and standard deviation over the five pursuer starts. The control interval is $\Delta_t = 0.1$~s. A single pursuer moves at a constant speed below the evader's maximum, either toward the evader (pure pursuit) or toward its predicted position (lead pursuit). An episode ends when the evader reaches the goal, is captured, collides with a static obstacle, reaches the maximum episode steps (75), or has no certified action available.

\begin{figure}[t]
      \centering
      \includegraphics[width=\columnwidth]{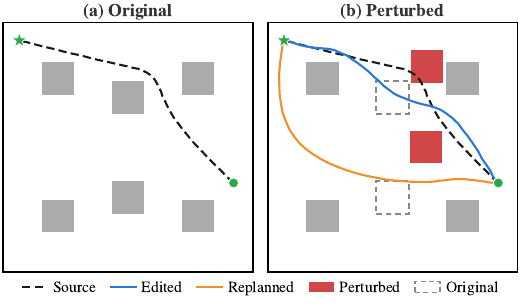}
    \vspace{-5mm}
    \caption{Editing vs.\ replanning after two obstacles move. (a)~Source plan on the original layout. (b)~The edited plan stays close to the source, while replanning takes a different route. Circle: start; star: goal.}
      \label{fig:qualitative_fe_vs_fm}
\end{figure}

\paragraph{Metrics}
We use four per-episode outcome rates, \textit{Goal Reach Rate} : the evader reaches the goal region, \textit{Collision Rate}
: collision with a static obstacle, \textit{Capture Rate}: the pursuer catches the evader, and \textit{Abort Rate}: the method finds no safe plan or action. Aborts occur only for the generative planners: for RAMP and SFM, no collision-free plan is found, and for TrajEdit, no safe escape action exists. Timeout rates are not reported, as the five outcomes sum to 100\%. Computation time is the mean computation per control step. In the perturbation study, we report the failure rate, the percentage of batches with no collision-free trajectory.

\paragraph{Baselines} We compare against three categories of planners. Classical: velocity obstacles (VO)~\cite{fiorini1998motion} and artificial potential fields (APF)~\cite{khatib1986real}. Learning-based: soft actor-critic (SAC)~\cite{haarnoja2018soft} and TD3~\cite{fujimoto2018addressing}. Generative: RAMP~\cite{teshome2025real}, which combines energy-based diffusion with potential fields, and SafeFlowMatcher (SFM)~\cite{yang2026safeflowmatcher}, which applies a CBF-QP during flow-matching generation.
We evaluate SFM in two variants: per-step, which regenerates at every step, and gated, which regenerates only when triggered by the same proximity gate as TrajEdit.

\begin{table*}[t]
\centering
\caption{Comparison with baselines under lead and pure pursuit on the convex layout. Best in bold, second best underlined (Goal and Cap.), -- : not applicable}
\label{tab:main_results}
\begin{threeparttable}
\footnotesize
\setlength{\tabcolsep}{3pt}
\begin{tabular}{l cccc cccc c}
\toprule
 & \multicolumn{4}{c}{Lead pursuit} & \multicolumn{4}{c}{Pure pursuit} & Time (ms)$\downarrow$ \\
\cmidrule(lr){2-5} \cmidrule(lr){6-9} \cmidrule(lr){10-10}
Method & Goal(\%) $\uparrow$ & Cap.(\%) $\downarrow$ & Coll.(\%) $\downarrow$ & Abort(\%) $\downarrow$ & Goal(\%) $\uparrow$ & Cap.(\%) $\downarrow$ & Coll.(\%) $\downarrow$ & Abort(\%) $\downarrow$ & Lead / Pure \\
\midrule
VO~\cite{fiorini1998motion} & $49.40\pm3.88$ & $50.60\pm3.88$ & $0.00\pm0.00$ & -- & $\bestm{99.00}\pm0.00$ & $\secondm{1.00}\pm0.00$ & $0.00\pm0.00$ & -- & 0.68 / 0.68 \\
APF~\cite{khatib1986real} & $5.00\pm0.63$ & $92.00\pm1.79$ & $0.00\pm0.00$ & -- & $74.40\pm3.20$ & $13.80\pm2.32$ & $0.00\pm0.00$ & -- & 0.13 / 0.13 \\
TD3~\cite{fujimoto2018addressing} & $54.20\pm1.94$ & $18.40\pm4.22$ & $23.80\pm2.14$ & -- & $85.20\pm4.45$ & $5.40\pm2.65$ & $9.00\pm2.19$ & -- & 0.25 / 0.26 \\
SAC~\cite{haarnoja2018soft} & $\secondm{57.00}\pm3.03$ & $\secondm{16.80}\pm2.32$ & $25.20\pm2.48$ & -- & $84.80\pm2.14$ & $5.40\pm1.85$ & $9.60\pm2.33$ & -- & 0.63 / 0.63 \\
RAMP~\cite{teshome2025real} & $33.40\pm3.50$ & $50.20\pm5.04$ & $0.00\pm0.00$ & $15.00\pm2.76$ & $70.40\pm3.61$ & $25.20\pm3.31$ & $0.00\pm0.00$ & $4.40\pm1.36$ & 528.87 / 495.85 \\
SFM~\cite{yang2026safeflowmatcher} (per-step) & $38.20\pm4.83$ & $25.40\pm1.96$ & $0.00\pm0.00$ & $35.00\pm5.37$ & $61.00\pm3.63$ & $19.20\pm2.48$ & $0.00\pm0.00$ & $18.20\pm1.33$ & 277.68 / 277.69 \\
SFM~\cite{yang2026safeflowmatcher} (gated) & $41.80\pm2.56$ & $ 23.80\pm5.19$ & $0.00\pm0.00$ & $32.40\pm3.61$ & $ 68.60\pm5.43$ & $17.60\pm5.85$ & $0.00\pm0.00$ & $12.80\pm2.48$ & 222.89/  228.92 \\
\midrule
TrajEdit (Ours) & $\bestm{83.00}\pm2.76$ & $\bestm{0.00}\pm0.00$ & $0.00\pm0.00$ & $ 16.00\pm3.22$ & $\secondm{96.80}\pm1.47$ & $\bestm{0.00}\pm0.00$ & $0.00\pm0.00$ & $ 2.80\pm1.72$ & 53.65 / 47.11 \\
\bottomrule
\end{tabular}
\end{threeparttable}
\end{table*}

\paragraph{Implementation}
Following the flow-matching implementation of~\cite{zhang2026affordance}, the velocity field is parameterized by a 1-D temporal U-Net~\cite{janner2022planning}. The network is conditioned via FiLM~\cite{perez2018film} on the concatenation of a 64-dimensional start-goal embedding, computed from the start and goal states by a three-layer MLP, and a 64-dimensional obstacle embedding, computed from the obstacle centers by a 1-D vision transformer. Both encoders are trained jointly with the flow model. To support variable-length plans with a single model, the horizon is sampled per batch from $\{4, 8, \dots, 48\}$. Conditioning dropout ($p = 0.2$), applied independently to the two embeddings, enables classifier-free guidance~\cite{ho2022classifier}. We optimize with AdamW at a learning rate of $10^{-4}$ and a batch size of 128 for 2{,}500 epochs, on an NVIDIA RTX 2080 Ti GPU. At inference, trajectories are generated with 10 steps and a guidance scale of 6, drawing 20 candidates per call, with the horizon adapted to the remaining distance to the goal. FlowEdit~\cite{kulikov2025flowedit} runs for 10 steps with source and target guidance scales of 3.0 and 4.5 and $n_{\max} = 4$. The CBF QP uses $\gamma = 4.3$ and is solved with CVXPY~\cite{diamond2016cvxpy, agrawal2018rewriting}. For the obstacle-perturbation study, each of 48 start-goal pairs in the convex layout has one source trajectory on the original layout and 40 perturbation draws per radius, displacing one or two obstacles; FlowEdit and FM each sample five candidates at a fixed horizon of 48. We implement SFM~\cite{yang2026safeflowmatcher} within our framework, using the same trained flow model as TrajEdit, with $T^p = 1$ prediction step and $T^c = 32$ correction steps. Its pursuer barrier is inflated by one pursuer step (Table~\ref{tab:sfm-sensitivity}).

\subsection{Results}
Table~\ref{tab:main_results} reports performance over 500 episodes per pursuit strategy (100 start-goal pairs, five initial pursuer positions). Under lead pursuit, TrajEdit reaches the goal in 83\% of episodes, 26.0 percentage points above the second best (SAC, 57.0\%). Under pure pursuit, it reaches the goal in 96.8\% of episodes, second only to VO (99.0\%), whose constant-velocity model of the pursuer closely matches pure pursuit; under lead pursuit, however, VO is captured in 50.6\% of episodes. Our method has 0\% capture rate and static collision under both pursuer strategies. Its remaining episodes are aborts, in which no safe escape action exists (16\% and 2.8\%) , and timeouts (1.0\% and 0.4\%). TD3 and SAC collide with static obstacles in 9--25\% of episodes, and RAMP and SFM are captured in 18--50\%. TrajEdit also requires 4--11$\times$ less computation per control step than RAMP and SFM.

Fig.~\ref{fig:qualitative_fe_vs_fm} shows the difference between editing and replanning. After two obstacles move, the source plan collides. Editing changes the plan only near the moved obstacles and keeps the rest of the source, while replanning from scratch produces a different route. Both plans are collision-free, but only the edited plan stays consistent with the plan the evader was already following.
\begin{table}
\centering
\caption{Ablation on the convex layout}
\label{tab:ablation_open}
\begin{threeparttable}
\footnotesize
\setlength{\tabcolsep}{4pt}
\begin{tabular}{l ccc}
\toprule
Variant & Goal(\%) $\uparrow$ & Abort(\%) $\downarrow$ & Time (ms) $\downarrow$ \\
\midrule
\multicolumn{4}{l}{\textit{Lead pursuit}} \\
\quad TrajEdit     & $83.00\pm2.76$ & $ 16.00\pm3.22$ & $ 53.65$ \\
\quad w/o FlowEdit & $81.80\pm3.54$ & $17.20\pm3.31$ & $ 65.63$ \\
\quad w/o planner     & $67.60\pm5.54$ & $27.20\pm5.23$ & $1.92$ \\
\midrule
\multicolumn{4}{l}{\textit{Pure pursuit}} \\
\quad TrajEdit     & $96.80\pm1.47$ & $2.80\pm1.72$ & $ 47.11$ \\
\quad w/o FlowEdit & $ 95.60\pm1.50$ & $4.40\pm1.50$ & $  64.33$ \\
\quad w/o planner     & $95.80\pm3.66$ & $4.20\pm3.66$ & $1.95$ \\
\bottomrule
\end{tabular}
\end{threeparttable}
\end{table}

\begin{table}
\centering
\caption{Ablation on the concave layout}
\label{tab:ablation_concave}
\begin{threeparttable}
\footnotesize
\setlength{\tabcolsep}{4pt}
\begin{tabular}{l ccc}
\toprule
Variant & Goal(\%) $\uparrow$ & Abort(\%) $\downarrow$ & Time (ms) $\downarrow$ \\
\midrule
\multicolumn{4}{l}{\textit{Lead pursuit}} \\
\quad TrajEdit     & $ 51.60\pm6.18$ & $43.60\pm3.98$& $54.54$ \\
\quad w/o FlowEdit & $ 55.80\pm4.26$ & $  38.00\pm2.76$ & $66.75$ \\
\quad w/o planner     & $ 49.80\pm5.15$ & $44.80\pm4.53$ & $1.95$ \\
\midrule
\multicolumn{4}{l}{\textit{Pure pursuit}} \\
\quad TrajEdit     & $ 75.80\pm2.79$ & $ 24.00\pm2.53$ &$  48.52$  \\
\quad w/o FlowEdit & $  73.40\pm2.42$ & $ 26.40\pm2.58$ & $ 65.28$ \\
\quad w/o planner     & $54.20\pm1.47$ & $45.80\pm1.47$ & $1.94$ \\
\bottomrule
\end{tabular}
\end{threeparttable}
\end{table}

 Tables~\ref{tab:ablation_open} and~\ref{tab:ablation_concave} show ablations of the adaptation mechanism under different layouts and pursuer strategies. Against full replanning (w/o FlowEdit) at the same trigger condition, editing reduces computation per control step by 18--27\% while matching its goal reach rate in three of the four settings. The exception is the concave layout under lead pursuit, where full replanning reaches the goal more often (55.8\% vs.\ 51.6\%). Without the planner (w/o planner), the CBF filter alone reaches the goal less often in all layouts and pursuit strategies, by 1.0--21.6 percentage points. No variant is captured or collides with an obstacle.

\begin{figure}[!h]
\centering
\includegraphics[width=\linewidth]{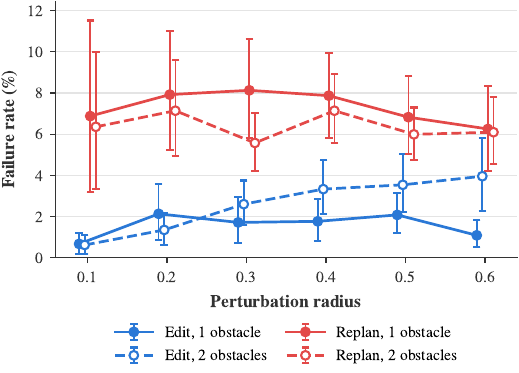}
\vspace{-5mm}
\caption{Failure rate under local perturbations of one and two static obstacles, across perturbation radius.}
\label{fig:ablation_radii}
\end{figure}

In Fig.~\ref{fig:ablation_radii}, we perturb one or two of the six obstacles of the convex layout within a given radius and run static inference. If none of the $N_c = 5$ sampled trajectories is collision-free, the run is counted as a planning failure. FlowEdit fails in 0.6--4.0\% of runs versus 5.6--8.1\% for flow matching from scratch, while being $1.7\times$ faster (33 vs.\ 56\,ms).

Table~\ref{tab:sfm-sensitivity} shows that, on the convex layout, SFM is captured in most episodes unless its barrier includes the pursuer's motion. Inflating the barrier by one pursuer step cuts captures to 17.6--19.2\% and raises the goal rate to 61.0--68.6\%; the speed-bound term in our CBF filter serves the same purpose. Gating regeneration with our proximity trigger further improves SFM, but even its best configuration remains well below TrajEdit (68.6\% vs.\ 96.8\% goal reach, 17.6\% vs.\ 0\% capture).

\begin{table}[!h]
\centering
\begin{threeparttable}
\caption{Sensitivity of SFM to the pursuer barrier radius (pure pursuit)}
\label{tab:sfm-sensitivity}
\begin{tabular}{lccc}
\toprule
  & Goal (\%) & Abort (\%) & Cap. (\%)  \\
\midrule
Per-step, $r_{\mathrm{cap}}$                      & $13.20 \pm 1.60$ & $4.60 \pm 2.24$  & $82.20 \pm 3.25$  \\
Gated, $r_{\mathrm{cap}}$                         & $18.00\pm3.03$ & $3.80\pm2.04$  & $ 78.20\pm3.19$ \\
Per-step, $r_{\mathrm{cap}} + v^p_{\max}\Delta_t$  & $61.00 \pm 3.63$ & $18.20 \pm 1.33$ & $19.20 \pm 2.48$  \\
Gated, $r_{\mathrm{cap}} + v^p_{\max}\Delta_t$     & $ 68.60\pm5.43$ & $ 12.80\pm2.48$ & $17.60\pm5.85$ \\
\bottomrule
\end{tabular}
\end{threeparttable}
\end{table}

\section{Conclusions}
In this paper, we presented TrajEdit, a trajectory generation framework for pursuit-evasion in dynamic environments. Instead of regenerating a plan whenever the environment changes, TrajEdit adapts the current plan to the new conditions, and uses a control barrier function filter when a moving obstacle comes within a safety threshold. In our experiments, TrajEdit recorded no captures or collisions in any evaluated setting, and editing reduced computation per control step by 18--27\% relative to full replanning at comparable success.








\FloatBarrier
\bibliographystyle{IEEEtran}
\bibliography{reference}

\end{document}